\pdfoutput=1

\documentclass[11pt]{article}

\usepackage{ACL2023}

\usepackage{times}
\usepackage{latexsym}

\usepackage[T1]{fontenc}

\usepackage[utf8]{inputenc}

\usepackage{microtype}

\usepackage{inconsolata}
\usepackage{microtype}

\usepackage{amsmath,amssymb,amsfonts}
\usepackage{algorithmic}
\usepackage{graphicx}
\usepackage{textcomp}
\usepackage{hyperref}
\usepackage{xcolor}
\usepackage{amsmath}
\usepackage{amssymb}
\usepackage{booktabs}
\usepackage{stfloats}
\usepackage{wrapfig}
\usepackage{multirow}
\usepackage{enumitem}
\usepackage{lipsum}

\usepackage{algorithm}
\usepackage{algorithmic}
\newcommand{\dataset}{IndoRobusta}

\setlist{topsep=1pt,itemsep=1pt,partopsep=1pt, parsep=1pt}

\title{{\dataset}: Towards Robustness Against Diverse \\Code-Mixed Indonesian Local Languages}

\author{Muhammad Farid Adilazuarda$^1$, Samuel Cahyawijaya$^3$, Genta Indra Winata$^2$, \\
\textbf{Pascale Fung$^3$}, \textbf{Ayu Purwarianti}$^1$ \\
  $^1$Institut Teknologi Bandung \quad $^2$Bloomberg \\
  $^3$The Hong Kong University of Science and Technology \\
  \texttt{faridlazuarda@gmail.com}}

\begin{document}
\maketitle

\begin{abstract}

Significant progress has been made on Indonesian NLP. Nevertheless, exploration of the code-mixing phenomenon in Indonesian is limited, despite many languages being frequently mixed with Indonesian in daily conversation. In this work, we explore code-mixing in Indonesian with four embedded languages, i.e., English, Sundanese, Javanese, and Malay; and introduce \texttt{IndoRobusta}, a framework to evaluate and improve the code-mixing robustness. Our analysis shows that the pre-training corpus bias affects the model's ability to better handle Indonesian-English code-mixing when compared to other local languages, despite having higher language diversity.




\end{abstract}

\section{Introduction}










Recent developments in Indonesian Natural Language Processing (NLP) have introduced an immense improvement in many aspects, including standardized benchmarks~\cite{wilie2020indonlu,cahyawijaya2021indonlg, koto2020indolem,winata2022nusax}, large pre-trained language model (LM)~\cite{wilie2020indonlu,cahyawijaya2021indonlg,koto2020indolem}, and resource expansion covering local Indonesian languages~\cite{apriani2016bugis,dewi2020madura,khaikal2021dayak}.
Despite all these significant efforts, only a few studies focus on tackling the code-mixing phenomenon that naturally occurs in the Indonesian language. Code-mixing
\footnote{In our case, code-mixing refers to intra-sentential code-switching where the language alternation occurs in the sentence.}
is an interesting phenomenon where people change between languages and mix them in a conversation or sentence. In Indonesia, many people speak at least two languages (i.e., Indonesian and a local language) in their day-to-day conversation~\cite{aji2022one}, and use diverse written and spoken styles specific to their home regions.



%


Inspired by the frequently occurring code-mixing phenomenon in Indonesian, we want to answer two research questions "\textit{Is the LMs performance susceptible to linguistically diverse Indonesian code-mixed text?}" and "\textit{How can we improve the model's robustness against a variety of mixed-language texts?}". Therefore, we introduce \texttt{IndoRobusta}, a framework to assess and improve code-mixed robustness. Using our \texttt{IndoRobusta-Blend}, we conduct experiments to evaluate existing pre-trained LMs using code-mixed language scenario to simulate the code-mixing phenomenon. We focus on Indonesian as the matrix language (L1) and the local language as the embedded language (L2)~\cite{myersscotton2009}.We measure the robustness of Indonesian code-mixed sentences for English (en) and three local languages, i.e, Sundanese (su), Javanese (jv), and Malay (ms)\footnote{Malay is not a direct Indonesian local language, but it is considered as the parent language to many of Indonesian local languages such as Jambi, Malay, Minangkabau, and Betawi.} on sentiment and emotion classification tasks. 
In addition, we explore methods to improve the robustness of LMs to code-mixed text. Using our \texttt{IndoRobusta-Shot}, we perform adversarial training to improve the code-mixed robustness of LMs. We explore three kinds of tuning strategies: 1) code-mix only, 2) two-steps, and 3) joint training, and empirically search for the best strategy to improve the model robustness on code-mixed data. 


We summarize our contribution as follows:
\begin{itemize}
\item We develop a benchmark to assess the robustness of monolingual and multilingual LMs on four L2 code-mixed languages covering English (en), Sundanese (su), Javanese (jv), and Malay (ms);
\item We introduce various adversarial tuning strategies to better improve the code-mixing robustness of LMs. Our best strategy improves the accuracy by $\sim$5\%  on the code-mixed test set and $\sim$2\% on the monolingual test set;
\item We show that existing LMs are more robust to English code-mixing rather than to local languages code-mixing and provide detailed analysis of this phenomenon.

\end{itemize}

\section{IndoRobusta Framework}

IndoRobusta is a code-mixing robustness framework consisting of two main modules: 1) \texttt{IndoRobusta-Blend}, which evaluates the code-mixing robustness of LMs through a code-mixing perturbation method, and 2) \texttt{IndoRobusta-Shot}, which improves the code-mixing robustness of LMs using a code-mixing adversarial training technique.

\subsection{Notation}
Given a monolingual language sentence $X = \{w_1, w_2, \dots, w_M\}$, where $w_i$ denotes a token in a sentence and $M$ denotes the number of tokens in a sentence, we denote a monolingual language dataset $\mathcal{D} = \{(X_1, Y_1), (X_2, Y_2), \dots, (X_N, Y_N)\}$, where $(X_i, Y_i)$ denotes a sentence-label pair and $N$ is the number of samples. Given a token $w_i$, a mask token $w^{mask}$ and a sentence $X$, we define a sentence with masked $w_i$ token as $X_{\backslash w_i} = \{w_1, w_2, \dots, w_{i-1}, w^{mask}, w_{i+1}, \dots, w_M\}$. We further define a code-mixing dataset $\mathcal{D}' = \{(X'_1, Y_1), (X'_2, Y_2), \dots, (X'_N, Y_N)\}$ where $X'_i$ denotes the code-mixed sentence. 
Lastly, we define  the set of parameters of a language model as $\theta$, the prediction label of a sentence $X$ as $f_\theta(X)$, the prediction score of the label $Y$ given a sentence $X$ as $f_\theta(Y|X)$, and the prediction score of the label other than $Y$ given a sentence $X$ as $f_\theta(\bar{Y}|X)$.

\subsection{IndoRobusta-Blend}
\label{sec:indorobusta-blend}


\texttt{IndoRobusta-Blend} is a code-mixing robustness evaluation  method that involves two steps: 1) code-mixed dataset generation and 2) model evaluation on the code-mixed dataset. The first step is synthetically generating the code-mixed example using the translation of important words in a sentence. 
To do so, we formally define the importance $I_{w_i}$ of the word $w_i$ for a given sample $(X, Y)$ as:
\[
I_{w_i} =  
    \begin{cases}
        f_\theta(Y|X) - f_\theta(Y|X_{\backslash w_i}), \\
        \qquad \text{if} f_\theta(X) = f_\theta(X_{\backslash w_i}) = Y\\
        [f_\theta(Y|X) - f_\theta(Y|X_{\backslash w_i})] + \\
        \qquad [f_\theta(\bar{Y}|X) - f_\theta(\bar{Y}|X_{\backslash w_i})], \text{otherwise}.
    \end{cases}
\]
\texttt{IndoRobusta-Blend} takes $R$\% words with the highest $I_{w_i}$, denoted as the \textbf{perturbation ratio}, and applies a word-level translation for each word. Using the translated words, \texttt{IndoRobusta-Blend} generates a code-mixed sentence by replacing the important words with their corresponding translation. To ensure generating a semantically-related code-mixed samples, we define a similarity threshold $\alpha$ to constraint the cosine distance between $X$ and $X_{adv}$. When the distance between $X$ and $X_{adv}$ is below $\alpha$, we resample the perturbed words and generate a more similar $X_{adv}$.

More formally, 
we define the code-mixing sample generation as a function $g(X,Y,\theta) = X_{adv}$. 
To generate the code-mixed dataset $\mathcal{D'}$ from the monolingual dataset $\mathcal{D}$ and a model $\theta$, \texttt{IndoRobusta-Blend} applies $g(X_i,Y_i,\theta)$ to each sample $(X_i, Y_i)$ in $\mathcal{D}$.
Using $\mathcal{D}$ and $\mathcal{D'}$, \texttt{IndoRobusta-Blend} evaluates the robustness of the fine-tuned model $\theta'$, trained on $\mathcal{D}$, by evaluating $\theta$ on both $\mathcal{D}$ and $\mathcal{D'}$. 
More formally, we define the code-mixed sample generation in Algorithm~\ref{alg:code-mixed-generation}. 
\begin{algorithm}[!t]
\caption{ Code-mixed sample generation workflow in IndoRobusta framework}
\label{alg:code-mixed-generation}
\begin{algorithmic} 

\REQUIRE Clean sentence example $X$, ground truth label $Y$, language model $\Theta$, similarity threshold $\alpha$, perturb ratio $R$, embedded Language $L$ 

\ENSURE Adversarial Example $X_{adv}$

\break

\STATE $Y' \leftarrow$ \textsc{Predict}($\Theta$, $X$)

\IF{$Y' \neq Y$}
\RETURN $X$
\ENDIF

\STATE $W \leftarrow R\%$ highest $I_{w_{i}}$ words in $X$
\STATE $W^L \leftarrow$ \textsc{Translate}($W$, target-language=$L$)
\STATE $X_{adv} \leftarrow$ \textsc{Perturb(X, $W^L$)}
\IF{\textsc{Sim}($X$, $X_{adv}$) < $\alpha$}
\WHILE{\textsc{Sim}($X$, $X_{adv}$) < $\alpha$}
\STATE $W^L \leftarrow \textsc{Resample}(W^L, I_{w_{i}})$
\STATE $X_{adv} \leftarrow \textsc{Perturb}(X, W^L)$
\ENDWHILE

\ENDIF
\RETURN $X_{adv}$
\end{algorithmic}
\end{algorithm}

\subsection{IndoRobusta-Shot}
\label{sec:indorobusta-shot}

\texttt{IndoRobusta-Shot} is a code-mixing adversarial defense  method, which aims to improve the robustness of the model. \texttt{IndoRobusta-Shot} does so by fine-tuning the model on the generated code-mixed dataset $\mathcal{D'}$. Similar to \texttt{IndoRobusta-Blend}, our \texttt{IndoRobusta-Shot} generates $\mathcal{D'}$ from $\mathcal{D}$ and $\theta$ by utilizing the code-mixed sample generation method $g(\theta,X,Y)$. Three different fine-tuning scenarios are explored in \texttt{IndoRobusta-Shot}
, i.e., \textbf{code-mixed-only tuning}, which fine-tune the model only on $\mathcal{D'}$; \textbf{two-step tuning}, which first fine-tune the model on $\mathcal{D}$, followed by a second-phase fine-tuning on $\mathcal{D'}$; and \textbf{joint training}, which fine-tunes the model on a combined dataset from $\mathcal{D}$ and $\mathcal{D'}$.









\section{Experiment Setting}

\subsection{Dataset}

We employ two Indonesian multi-class classification datasets for conducting our experiments, i.e., a sentiment-analysis dataset, SmSA~\cite{purwarianti2019improving}, and an emotion classification dataset, EmoT~\cite{saputri2018emotion}. SmSA is a sentence-level sentiment analysis dataset 
consists of 12,760 samples and is labelled intro three possible sentiments values, i.e., positive, negative, and neutral. EmoT is an emotion classification dataset 
which consists of 4,403 samples and covers five different emotion labels, i.e., anger, fear, happiness, love, and sadness. The statistics of SmSA and EmoT datasets are shown in Appendix Table~\ref{tab:dataset_statistics}.

\begin{table}[t!]
  \centering
  \resizebox{\linewidth}{!}{
  \begin{tabular}{lcccccc}
    \toprule
        {\textbf{Model}} & {\textbf{Orig.}} & {\textbf{en}} & {\textbf{jw}} & {\textbf{ms}} & {\textbf{su}} & \textbf{avg} \\
        \midrule
        \multicolumn{7}{c}{EmoT} \\ \midrule
        IB\textsubscript{\textsc{B}} & 72.42 & \underline{9.55} & \underline{12.35}  & \textbf{9.47} & \underline{9.39} &  \textbf{10.19} \\
        IB\textsubscript{\textsc{L}} &  \underline{75.53} & \textbf{9.24} &  \textbf{12.12} & \underline{10.23} & \textbf{9.32} &  \underline{10.23} \\
        mB\textsubscript{\textsc{B}} & 61.14 & 12.50 & 14.02 & 12.73 & 12.50 & 12.96 \\
        XR\textsubscript{\textsc{B}} & 72.88 & 10.98 & 13.94 & 13.18 & 12.50 &  12.65 \\
        XR\textsubscript{\textsc{L}} & \textbf{78.26} & 12.27 & 13.03 & 12.42 & 11.74 & 12.37 \\
        \midrule
        Avg & & 10.91 & 13.09 & 11.61 &  11.09 & \\
        
        \midrule
        
        \multicolumn{7}{c}{SmSA} \\ \midrule
        IB\textsubscript{\textsc{B}} & 91.00 & \textbf{1.33} & 5.07 & 3.20 & \underline{2.40}  & 3.00 \\
        IB\textsubscript{\textsc{L}} & \textbf{94.20} & 2.47 & 4.13 & 4.00 & \textbf{2.20} & 3.20 \\
        mB\textsubscript{\textsc{B}} & 83.00 & 2.20 & \textbf{3.00} & \underline{2.93} & 2.47 &  \textbf{2.65} \\
        XR\textsubscript{\textsc{B}} & 91.53 & 3.40 & 3.80 & 4.27 & 4.27 & 3.94 \\
        XR\textsubscript{\textsc{L}} &  \underline{94.07} & \underline{2.13} & \underline{3.20} & \textbf{2.60} & 2.73 & \underline{2.67} \\
        \midrule
        Avg & & 2.31 & 3.84 &3.40 & 2.81 &  \\
    \bottomrule
  \end{tabular}
  }
  \caption{Delta accuracy with $R=0.4$ on the test data. A lower value denotes better performance. We \textbf{bold} the best score and \underline{underline} the second-best score.}
  \label{tabel:robustness_result}
\end{table}

\subsection{Code-mixed Sample Generation}

For our experiment, we use Indonesian as the L1 language and explore four commonly used L2 languages, i.e., English, Sundanese, Javanese, and Malay.
We experiment with different code-mixed perturbation ratio $R = \{0.2, 0.4, 0.6, 0.8\}$ to assess the susceptibility of models. We utilize Google Translate to translate important words to generate the code-mixed sentence $X'$.




\subsection{Baseline Models}
We include both monolingual and multilingual pre-trained LMs with various model size in our experiment.
For Indonesian monolingual pre-trained LMs, we utilize two models: IndoBERT\textsubscript{\textsc{BASE}} (IB\textsubscript{\textsc{B}}) and IndoBERT\textsubscript{\textsc{LARGE}} (IB\textsubscript{\textsc{L}})~\cite{wilie2020indonlu}, while for the multilingual LMs, we employ mBERT\textsubscript{\textsc{BASE}} (mB\textsubscript{\textsc{B}})~\cite{devlin-etal-2019-bert}, XLM-R\textsubscript{\textsc{BASE}} (XR\textsubscript{\textsc{B}}), and XLM-R\textsubscript{\textsc{LARGE}} (IB\textsubscript{\textsc{L}})~\cite{conneau-etal-2020-unsupervised}. Note that all of the multilingual models are knowledgeable of the Indonesian language and all L2 languages used since all the languages are covered in their pre-training corpus.

\begin{table}[t!]
  \centering
  \resizebox{\linewidth}{!}{
    \begin{tabular}{lcccccc}
    \toprule
        \textbf{Model} & \multicolumn{2}{c}{\textbf{CM Only}} & \multicolumn{2}{c}{\textbf{Two-Step}} & \multicolumn{2}{c}{\textbf{Joint}} \\
        & \textbf{Orig} & \textbf{CM}
        & \textbf{Orig} & \textbf{CM}
        & \textbf{Orig} & \textbf{CM} \\
        \midrule
        \multicolumn{7}{c}{EmoT} \\ \midrule
        IB\textsubscript{\textsc{B}} & 45.13 & 66.53 & 69.85 & 68.31 & 74.68 & 67.27 \\
        IB\textsubscript{\textsc{L}} & \underline{63.29} & \underline{68.58} & \underline{73.06} & \underline{69.46} & \underline{75.90} & \underline{68.01} \\
        mB\textsubscript{\textsc{B}} & 32.97 & 58.11 & 54.72 & 59.68 & 62.98 & 56.54 \\
        XR\textsubscript{\textsc{B}} & 57.59 & 68.40 & 72.17 & 69.11 & 74.38 & 67.26 \\
        XR\textsubscript{\textsc{L}} & \textbf{71.61} & \textbf{71.56} & \textbf{77.13} & \textbf{70.44} & \textbf{78.31} & \textbf{70.06} \\
        
        \midrule
        \multicolumn{7}{c}{SmSA} \\ \midrule
        IB\textsubscript{\textsc{B}} & 45.10 & 93.51 & \underline{89.81} & 92.68 & 92.52 & 90.71 \\
        IB\textsubscript{\textsc{L}} & \textbf{68.40} & \underline{94.67} & \textbf{90.60} & \underline{94.12} & \underline{94.73} & \underline{93.00} \\
        mB\textsubscript{\textsc{B}} & 51.72 & 83.73 & 78.95 & 85.16 & 85.61 & 84.31 \\
        XR\textsubscript{\textsc{B}} & 59.31 & 91.37 & 68.08 & 93.87 & 93.77 & 92.21 \\
        XR\textsubscript{\textsc{L}} & \underline{63.06} & \textbf{95.07} & 85.96 & \textbf{95.35} & \textbf{95.35} & \textbf{93.99} \\
        
        
        \bottomrule
    \end{tabular}
  }
  \caption{Accuracy on original (Orig.) and code-mixing (CM) test sets after adversarial training with different tuning strategies.}
  \label{tabel:adv_train_on_perturbed_test_set}
\end{table}

\subsection{Training Setup}

To evaluate the model robustness,
We fine-tune the model on $D$ using the Adam optimizer~\cite{kingma2014adam} with a learning rate of 3e-6, and a batch size of 32. We train the model for a fixed number of epoch, i.e., 5 epochs for sentiment analysis and 10 epochs for emotion classification. 
We run each experiment three times using different random seeds and report the averaged score over three runs. For the adversarial training,
we train the model using Adam optimizer with a learning rate of 3e-6 and a batch size of 32. 
We set the maximum epoch to 15, and apply early stopping
with the early stopping patience set to 5. 




\subsection{Evaluation Setup}



To measure the robustness of the models, \texttt{IndoRobusta} uses three evaluation metrics: 1) the accuracy on the monolingual dataset, 2) the accuracy on the code-mixed dataset, and 3) delta accuracy~\cite{srinivasan2018delta}. We measure accuracy before and after adversarial training to analyze the effectiveness of the adversarial training method in the \texttt{IndoRobusta-Shot}.

\begin{figure}[t!]
    \centerline{\includegraphics[width=\columnwidth]{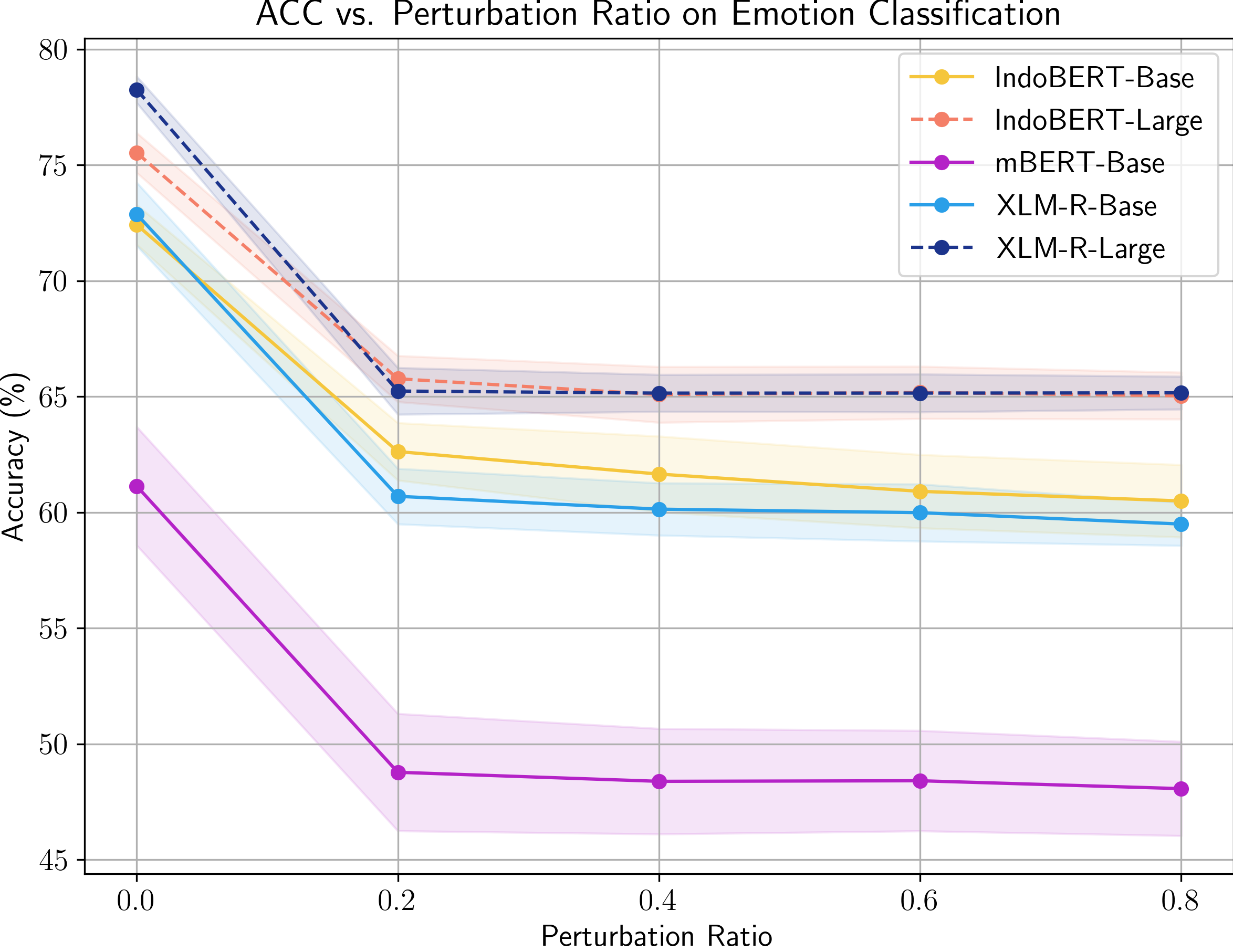}}
    \caption{The effect of perturbation ratio to the evaluation accuracy in the emotion classification task.}
    \label{fig:acc_vs_pert_emotion}
\end{figure}

\section{Result and Discussion}

\subsection{Code-Mixing Robustness}

The result of the robustness evaluation with $R=0.4$ is shown in Table~\ref{tabel:robustness_result}. Existing LMs are more prone to code-mixing in the emotion classification task,  with $>10\%$ performance reduction, compared to $~3\%$ on the sentiment analysis task. Interestingly, monolingual models, i.e., IndoBERT\textsubscript{\textsc{BASE}} and IndoBERT\textsubscript{\textsc{LARGE}}, are more robust in the emotion classification task compared to the multilingual models with $~2\%$ higher delta accuracy. While on the sentiment analysis task, all models perform almost equally good in all L2 languages.

We also observe that the robustness on English language are generally lower than Javanese and Malay in all models. We conjecture that this is due to the bias from the pre-training corpus, since pre-training corpus is gathered from online platforms, and Indonesian-English code-mixing is particularly common in such platforms~\cite{bani2018enid,aulia2017enid,marzona2017enid}. While Indonesian and local language code-mixing are considered a secondary choice in online platforms~\cite{cahyani2020socmed} and is more commonly used in the day-to-day conversation ~\cite{ginting2019suid,muslimin2020jvid}.

\subsection{Impact of Perturbation Ratio}



According to Figure~\ref{fig:acc_vs_pert_emotion}, we can clearly observe that LMs performance gets lower as the perturbation ratio $R$ increases. Interestingly, the steepest decline happens when the perturbation ratio $R=0.4$, and the model performance decreases slightly with a higher perturbation ratio ($R=\{0.4, 0.6, 0.8\}$). This result suggests that translating the words with high importance as mentioned in \S\ref{sec:indorobusta-blend}, effectively alters the model prediction. 

We further analyzed the generated code-mixed sentence, we show the example of the generated code-mixed sentences from ~\texttt{IndoRobusta} in Table~\ref{tab:example}. To generate the code-mixed sentence, we select important words from the sentence and perform word-level translation into four different L2 languages, i.e English, Sundanese, Javanese, and Malay. 
We analyze the important word selected by the $I_{w_i}$ over a dataset, we count the total number of times a word is selected as important with $R=\{0.2, 0.4, 0.6, 0.8\}$, denoted as informative frequency (IF).
For each word, we divide the IF with its document frequency (DF) to produce a normalized informative frequency (IF/DF). We show the top-20 words with highest IF/DF score for emotion classification task in Table~\ref{tabel:top_20_emot} and for sentiment analysis task in Table~\ref{tabel:top_20_smsa}. Most of the words are related to the label in the lexical-sense, e.g.: 'regret', 'disappointing', and 'disappointed' are commonly associated with \textbf{negative} sentiment, while 'comfortable', 'fun', 'nice' are commonly associated with \textbf{positive} sentiment. Most of the time, the word-translations for all L2 languages are valid  and infer similar meaning.
We find that the model prediction is still largely shifted even though the important word is translated correctly. This shows that, despite having learned all the languages individually, LMs are unable to generalize well on code-mixed sentences and improving robustness with an explicit tuning is required to achieve comparable performance.

\begin{table}[!t] 
    \centering
    \small
    \resizebox{0.49\textwidth}{!}{
        \begin{tabular}{@{}p{0.44\linewidth}  p{0.47\linewidth}@{}}
            \toprule
             \textbf{Code-Mixed Text} & \textbf{Translation} \\
             \midrule
            sate kambing dan gulai kambing nya \textbf{\textcolor{blue}{sedap}} penyajian makannan nya juga sangat cepat tempat nya cukup bersih & lamb satay and lamb curry are \textbf{yummy}, quick serving, and the place is quite clean \\ 
            \midrule
            \textbf{\textcolor{orange}{hayam}} goreng, tempe, tahu goreng dengan sambal yang pedas mantap sejak zaman dulu \textbf{\textcolor{orange}{teu}} dan terjangkau & fried \textbf{chicken}, tempe, fried tofu with spicy chilli sauce has been \textbf{delicious} since ancient times.\\
            \midrule
            tidak bisa \textbf{\textcolor{red}{mudhun}} galau mikirin lo & I cannot \textbf{sleep} because I am thinking about you \\ 
            \midrule
            meski masa kampanye sudah selesai bukan berati habis pula \textbf{\textcolor{violet}{effort}} mengerek tingkat kedipilihan elektabilitas. & Even though the campaign period is over, it doesn't mean that the \textbf{effort} to raise the electability level is over. \\
            
            \bottomrule
        \end{tabular}
    }
    \caption{Example of generated code-mixed sentences with ~\texttt{IndoRobusta}. \textcolor{blue}{\textbf{Blue}} denotes an Malay word, \textcolor{orange}{\textbf{Orange}} denotes a Sundanese word, \textcolor{red}{\textbf{Red}} denotes a Javanese word and \textcolor{violet}{\textbf{Violet}} denotes an English word. The \textbf{bold words} in the translation column are the corresponding colored word translations in English.}
    \label{tab:example}
\end{table}


\subsection{Improving Code-Mixing Robustness}

Table~\ref{tabel:adv_train_on_perturbed_test_set} shows the results of the adversarial training using different tuning strategies. \textbf{Code-mixing only} and \textbf{two-step}-tuning yield a better improvement on the code-mixed data compared to the \textbf{joint training}. Nevertheless, \textbf{code-mixing only}-tuning significantly hurts the performance on the original data, while the \textbf{two-step}-tuning can retain much better performance on the original data. \textbf{joint training}, on the other hand, yields the highest performance on the original data, and even outperforms the model trained only on the original data by $\sim2\%$ accuracy while maintaining considerably high performance on the code-mixing data.





\section{Related Work}

\paragraph{Code-Mixing in NLP}

Code-mixing has been studied in various language pairs such as Chinese-English~\cite{lyu2010seame,winata2019code,lin2021bitod,lovenia2022ascend}, Cantonese-English~\cite{dai2022ciavsr}, Hindi-English~\cite{banerjee2018dataset,khanuja2020gluecos}, Spanish-English~\cite{calcs2018shtask,winata2019learning,aguilar2020lince}, Indonesian-English~\cite{barik2019normalization,stymne2020evaluating}, Arabic-English~\cite{hamed2019code}, etc. Multiple methods have been proposed to better understand code-mixing including multi-task learning~\cite{song2017multitask,winata2018multitaskcs}, data augmentation~\cite{winata2019code,chang2019code,lee2019linguistically,qin2020cosdamlmc,jayanthi2021codemixednlp,rizvi2021gcm}, meta-learning~\cite{winata2020mtl}, and multilingual adaptation~\cite{winata2021multics}. In this work, we explore code-mixing in Indonesian with four commonly used L2 languages.

\paragraph{Model Robustness in NLP}

Prior works in robustness evaluation focus on data perturbation methods~\cite{tan2021code,ishii2022question}. Various textual perturbation methods have been introduced~\cite{jin2019textfooler,dhole2021nl}, which is an essential part of robustness evaluation. Moreover, numerous efforts in improving robustness have also been explored, including adversarial training on augmented data~\cite{li2021defender,li-specia-2019-improving}, harmful instance removal~\cite{bang2021covid,kobayashi2020efficient} and robust loss function~\cite{bang2021covid,zhang2018gce}. In this work, we focus on adversarial training, since the method is effective for handling low-resource data, such as code-mixing.






\section{Conclusion}
We introduce \texttt{IndoRobusta}, a framework to effectively evaluate and improve model robustness. Our results suggest adversarial training can significantly improve the code-mixing robustness of LMs, while at the same time, improving the performance on the monolingual data. Moreover, we show that existing LMs are more robust to English code-mixed and conjecture that this comes from the source bias in the existing pre-training corpora.

\section*{Limitations}
One of the limitation of our approach is that we utilize Google Translate to generate the perturbed code-mixing samples instead of manually generating natural code-mixing sentences. Common mistake made from the generated code-mixed sentence is on translating ambiguous terms, which produces inaccurate word-level translation and alters the meaning of the sentence. For future work, we expect to build a higher quality code-mixed sentences to better assess the code-mixed robustness of the existing Indonesian large-pretrained language models.

\section*{Acknowledgements}
We sincerely thank the anonymous reviewers for their insightful comments on our paper.


\bibliography{anthology,custom}
\bibliographystyle{acl_natbib}\clearpage

\appendix

\section{Annotation Guideline for Human Evaluation}
\label{app:human-annotation}


We introduce a manual annotation to evaluate the generated code-mixed sentences. To validate the quality of our perturbed code-mixing sentences, we hire 3 native annotators for each language to evaluate the generated Sundanese-Indonesian and Javanese-Indonesian code-mixed sentences, and 3 Indonesian annotators with professional English proficiency for assessing the generated English-Indonesian code-mixed sentences. Each human annotator is asked to assess the quality of 40 randomly sampled code-mixed sentences and provide a score in range of $[1,2,3,4,5]$ with 1 denotes an incomprehensible code-mixing sentence and 5 denotes a perfectly natural code-mixed sentence. The detailed annotation guideline is described in \ref{app:human-annotation} The score between annotators are averaged to reduce annotation bias.

\begin{table}[H]
\centering
\resizebox{0.9\linewidth}{!}{
\begin{tabular}{lcccc}
\toprule
\textbf{Dataset} & $|$\textbf{Train}$|$ & $|$\textbf{Valid}$|$ & $|$\textbf{Test}$|$  & \textbf{\#Class} \\ \midrule
EmoT & 3,521 & 440 & 442 & 5 \\ 
SmSA & 11,000 & 1,260 & 500 & 3 \\ 
\bottomrule
\end{tabular}
}
\caption{Statistics of EmoT and SmSA datasets.}
\label{tab:dataset_statistics}
\end{table}

Table \ref{tab:dataset_statistics} contains more details of the EmoT and SmSA dataset that we used in the sample generation. Sample generated by perturbing these datasets will later be annotated.

First, we compile 40 samples generated from each model into an excel sheet. Then the annotator is given access to the file. Before starting the annotation process, the annotator is given instructions and a definition of the score that can be assigned to the sample sentence. For each row in the given excel file, the annotator is asked to read the code-mixing sentence generated by the model and provide annotation values. Annotation scores are defined as follows:\\
\textbf{1 - unnatural} (unintelligible sentence) \\
\textbf{2 - less natural} (sentences can be understood even though they are strange) \\
\textbf{3 - adequately natural} (sentences can be understood even though they are not used correctly) \\
\textbf{4 - imperfect natural} (sentences are easy to understand, but some of the words used are slightly inaccurate) \\
\textbf{5 - natural} (sentences are easy to understand and appropriate to use) \\

\section{Annotation Result}
\begin{figure}[H]
    \centerline{\includegraphics[width=\columnwidth]{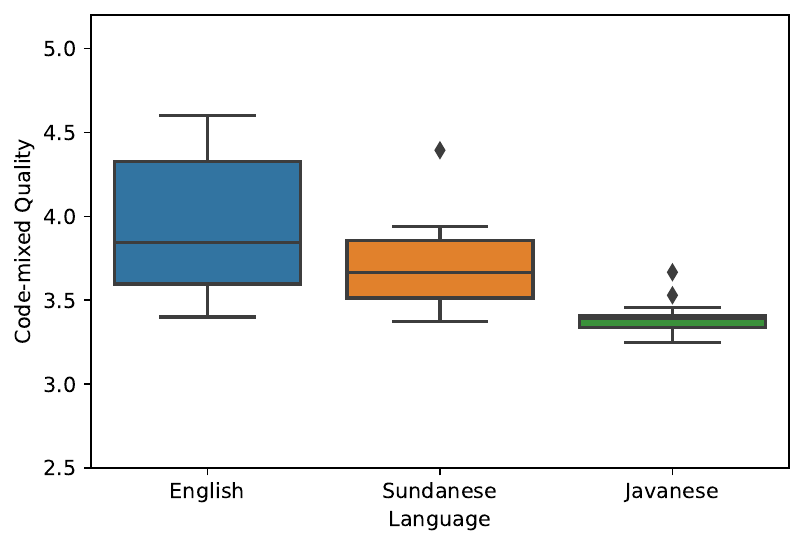}}
    \caption{Human evaluation result from the generated code-mixed samples averaged over three annotators.}
    \label{fig:human_evaluation}
\end{figure}
Figure~\ref{fig:human_evaluation} shows the result of the human assessment on the generated code-mixed sentences. The results indicates that the generated sentences are adequately natural by achieving an average score of 3.94 for English-Indonesian, 3.71 for Sundanese-Indonesian, and 3.39 for Javanese-Indonesian.



\begin{table*}[ht]
  \centering
  \resizebox{1\linewidth}{!}{
  \begin{tabular}{lccccccc}
    \toprule
    {\textbf{Word}} & {\textbf{IF}} & {\textbf{DF}} & {\textbf{IF/DF}} & {\textbf{jw}} & \textbf{ms} & \textbf{su} & \textbf{en} \\
    \midrule
    
    love & 1078 & 1260 & 0.856 & tresna & cinta & cinta & love \\
    tolong & 1408 & 2520 & 0.559 & bantuan & membantu & Tulung & help \\
    km & 1183 & 2520 & 0.469 & km & km & km & km \\
    kasih & 2947 & 6300 & 0.468 & tresna & cinta & cinta & love \\
    pakai & 1505 & 3360 & 0.448 & nggunakake & guna & ngagunakeun & use \\
    udh & 1659 & 3780 & 0.439 & wis & Sudah & Geus & Already \\
    setan & 1088 & 2520 & 0.432 & setan & syaitan & Sétan & Devil \\
    hrs & 1078 & 2520 & 0.428 & \textbf{\textcolor{red}{jam}} & \textbf{\textcolor{red}{jam}} & \textbf{\textcolor{red}{tabuh}} & \textbf{\textcolor{red}{hrs}} \\
    cinta & 5559 & 13020 & 0.427 & tresna & cinta & \textbf{\textcolor{red}{cinta}} & love \\
    jam & 2495 & 5880 & 0.424 & jam & pukul & tabuh & o'clock \\
    gua & 1594 & 3780 & 0.422 & aku & saya & abdi & I \\
    jatuh & 1768 & 4200 & 0.421 & tiba & jatuh & ragrag ka handap & fall down \\
    mobil & 1057 & 2520 & 0.419 & mobil & kereta & mobil & car \\
    sehat & 1214 & 2940 & 0.413 & \textbf{\textcolor{red}{sehat}} & sihat & cageur & healthy \\
    beneran & 1351 & 3360 & 0.402 & tenan & sungguh & saleresna & really \\
    kadang & 1175 & 2940 & 0.400 & kadhangkala & kadang-kadang & sakapeung & sometimes \\
    lu & 1505 & 3780 & 0.398 & \textbf{\textcolor{red}{lu}} & \textbf{\textcolor{red}{lu}} & \textbf{\textcolor{red}{lu}} & \textbf{\textcolor{red}{lu}} \\
    ketemu & 1641 & 4200 & 0.391 & ketemu & berjumpa & papanggih & meet \\
    dgn & 2254 & 5880 & 0.383 & karo & dengan & kalawan & with \\
    kantor & 1127 & 2940 & 0.383 & kantor & pejabat & kantor & office \\
        
    \bottomrule
  \end{tabular}
  }
  \caption{Top 20 most perturbed word on \textbf{emotion classification} experiments conducted on test data and their translation on four languages. \textbf{\textcolor{red}{Red}} denotes mistranslated words due to ambiguity or translator limitation.}
  \label{tabel:top_20_emot}
\end{table*}

\begin{table*}[ht]
  \centering
  \resizebox{1\linewidth}{!}{
  \begin{tabular}{lccccccc}
    \toprule
        {\textbf{Word}} & {\textbf{IF}} & {\textbf{DF}} & {\textbf{IF/DF}} & {\textbf{jw}} & \textbf{ms} & \textbf{su} & \textbf{en} \\
        \midrule
        cocok & 1750 & 2100 & 0.833 & cocok & sesuai & cocog & suitable \\
        asik & 2338 & 2940 & 0.795 & Asik & Asik & Asik & Asik \\
        nyaman & 2905 & 3780 & 0.769 & nyaman & selesa & sreg & comfortable \\
        menyesal & 2240 & 2940 & 0.76 & getun & penyesalan & kaduhung & regret \\
        mantap & 8456 & 11340 & 0.746 & ajeg & mantap & ajeg & steady \\
        mengecewakan & 3094 & 4200 & 0.737 & nguciwani & mengecewakan & nguciwakeun & disappointing \\
        kecewa & 21910 & 30660 & 0.715 & kuciwa & kecewa & kuciwa & disappointed \\
        enak & 9443 & 14700 & 0.642 & \textbf{\textcolor{red}{becik}} & bagus & hade & nice \\
        jelek & 1617 & 2520 & 0.642 & ala & teruk & goréng & bad \\
        salut & 1834 & 2940 & 0.624 & salam & tabik hormat & salam & salute \\
        memuaskan & 2877 & 4620 & 0.623 & marem & memuaskan & nyugemakeun & satisfying \\
        keren & 3136 & 5040 & 0.622 & \textbf{\textcolor{red}{kelangan}} & \textbf{\textcolor{red}{sejuk}} & tiis & cool \\
        kadaluarsa & 1827 & 2940 & 0.621 & kadaluarsa & tamat tempoh & kadaluwarsa & expired \\
        murah & 3094 & 5040 & 0.614 & murah & murah & murah & inexpensive \\
        kartu & 2058 & 3360 & 0.613 & kertu & kad & kartu & card \\
        banget & 2434 & 	41160 & 0.591 & banget & sangat & pisan & very \\
        bangga & 148 & 	2520 & 0.589 & bangga & bangga & reueus & proud \\
        mending & 1974 & 3360 & 0.588 & luwih apik & lebih baik & Leuwih alus & Better \\
        uang & 4396 & 7560 & 0.581 & dhuwit & wang & duit & money \\
        id & 1442 & 2520 & 0.572 & id & ID & \textbf{\textcolor{red}{en}} & id \\
        
    \bottomrule
  \end{tabular}
  }
  \caption{Top 20 most perturbed word on \textbf{sentiment analysis} experiments conducted on test data and their translation on four languages. \textbf{\textcolor{red}{Red}} denotes mistranslated words due to ambiguity or translator limitation.}
  \label{tabel:top_20_smsa}
\end{table*}

\end{document}